\documentclass{llncs}
\usepackage{llncsdoc,amsmath,graphicx}
\usepackage{algorithm}
\usepackage{mathtools}
\usepackage{amssymb}
\usepackage{amsmath}
\usepackage{graphicx}
\usepackage{subcaption}
\usepackage{algorithm}
\usepackage[noend]{algpseudocode}
\usepackage[misc]{ifsym}
\usepackage{lipsum}
\usepackage{tikz}
\usepackage{todonotes}

\newcommand\blfootnote[1]{%
  \begingroup
  \renewcommand\thefootnote{}\footnote{#1}%
  \addtocounter{footnote}{-1}%
  \endgroup
}

\renewcommand{\thefootnote}{\fnsymbol{footnote}}

\title{Event-Based Modeling with High-Dimensional Imaging Biomarkers for Estimating Spatial Progression of Dementia}

\begin{document}
%\ninept
\date{}
\author{}
\institute{}
\author{Vikram Venkatraghavan\textsuperscript{*} \inst{1} \textsuperscript{\Letter} \and
Florian Dubost\textsuperscript{*} \inst{1} \and
Esther  E. Bron \inst{1} \and
\mbox{Wiro J. Niessen \inst{1,2}} \and
Marleen de Bruijne \inst{1,3} \and
Stefan Klein \inst{1}, \mbox{for the Alzheimer's Disease Neuroimaging Initiative}}
\institute{$^1$ Biomedical Imaging Group Rotterdam, Depts. of Medical Informatics \& Radiology, Erasmus MC, The Netherlands \\
$^2$ Faculty of Applied Sciences, Delft University of Technology, The Netherlands \\
$^3$ Department of Computer Science, University of Copenhagen, Denmark \\
\email{v.venkatraghavan@erasmusmc.nl} }

\maketitle
\blfootnote{\textsuperscript{*} Contributed equally to the study }
\begin{abstract}

Event-based models (EBM) are a class of disease progression models that can be used to estimate temporal ordering of neuropathological changes from cross-sectional data. Current EBMs only handle scalar biomarkers, such as regional volumes, as inputs. However, regional aggregates are a crude summary of the underlying high-resolution images, potentially limiting the accuracy of EBM. Therefore, we propose a novel method that exploits high-dimensional voxel-wise imaging biomarkers: n-dimensional discriminative EBM (nDEBM). nDEBM is based on an insight that mixture modeling, which is a key element of conventional EBMs, can be replaced by a more scalable semi-supervised support vector machine (SVM) approach. This SVM is used to estimate the degree of abnormality of each region which is then used to obtain subject-specific disease progression patterns. These patterns are in turn used for estimating the mean ordering by fitting a generalized Mallows model. In order to validate the biomarker ordering obtained using nDEBM, we also present a framework for Simulation of Imaging Biomarkers' Temporal Evolution (SImBioTE) that mimics neurodegeneration in brain regions. SImBioTE trains variational auto-encoders (VAE) in different brain regions independently to simulate images at varying stages of disease progression. We also validate nDEBM clinically using data from the Alzheimer's Disease Neuroimaging Initiative (ADNI). In both experiments, nDEBM using high-dimensional features gave better performance than state-of-the-art EBM methods using regional volume biomarkers. This suggests that nDEBM is a promising approach for disease progression modeling.

\end{abstract}

\section{Introduction} \label{sec:intro}

In 2015, approximately 46.8 million people were estimated to be living with dementia, and by 2050 this number is expected to have increased to 131.5 million~\cite{Prince:2015}. Dementia is characterized by a cascade of neuropathological changes which are quantified using several imaging and non-imaging biomarkers. Understanding how the different biomarkers progress from normal to abnormal state after disease onset enables precise estimation of disease severity in an objective and quantitative way. This can help in identifying individuals at risk of developing dementia as well as monitor the effectiveness of preventive and supportive therapies.

Event-based models (EBM) are a class of disease progression models that estimate  the order in which biomarkers become abnormal during disease progression using cross-sectional data~\cite{Fonteijn:2012,Vikram:2019,Young:2014,Huang:2012}. It was reported in a recent paper on discriminative EBM (DEBM)~\cite{Vikram:2019} that the EBMs are very sensitive to the quality of biomarkers used for building the model. Hence, to infer the neuropathological changes that occur during dementia accurately, good quality biomarkers are important.

An essential step in an EBM involves mixture modeling to obtain biomarker distributions in normal and abnormal classes~\cite{Fonteijn:2012,Vikram:2019}. This restricts the current EBMs to only handle scalar biomarkers. In case of imaging biomarkers, regional volumes from structural MRIs are often used~\cite{Vikram:2019,Neil:2017,Young:2017,Young:2014,Fonteijn:2012}. However, regional volumes are a crude summary of the high-dimensional information available from structural MRI, resulting in suboptimal EBM performance, as shall be demonstrated later in this paper. Therefore, we propose a novel method  that exploits voxel-wise imaging biomarkers: n-dimensional discriminative EBM (nDEBM).

Estimating the accuracy of ordering obtained by EBMs is not feasible as ground-truth ordering is not known for a disease. In order to validate the proposed method and compare its accuracy with that of existing state-of-the-art EBM methods, we also present a framework for Simulation of Imaging Biomarkers' Temporal Evolution (SImBioTE).  SImBioTE uses variational auto-encoders (VAE) to simulate neurodegeneration in brain regions. These regions are represented by a vector in the latent space of the VAE. Synthetic brain regions were created by sampling latent representations corresponding to target degrees of abnormality which were determined by a ground-truth ordering of disease progression. The generated synthetic brain regions were used as inputs for nDEBM, and the regional aggregates were used as inputs for state-of-the-art EBMs to evaluate the accuracies.

\section{nDEBM} \label{sec:nDEBM}

In Section~\ref{ssec:debm}, a brief introduction to the current DEBM~\cite{Vikram:2019} model is given. Section~\ref{ssec:biopro}, presents a novel framework to use semi-supervised SVMs in DEBM for estimating posterior probabilities of abnormality for high-dimensional biomarkers. In Section~\ref{ssec:staging}, we use these posterior probabilities to estimate severity of disease progression in an individual.

\subsection{DEBM} \label{ssec:debm}

In a cross-sectional dementia dataset $(X)$ of $M$ subjects (consisting of cognitively normal (CN) and patients with dementia (DE)), let $X_j$ denote a measurement of biomarkers for subject $j\in\left[1,M\right]$, consisting of $N$ scalar biomarker values $x_{j,i}$. As dementia is characterized by a cascade of neuropathological changes that occurs over several years, even CN subjects can show some abnormal biomarker values. On the other hand, in DE subjects, a proportion of biomarkers may still have normal values, especialy in patients at an early disease stage. This leads to label noise in the data and hence clinical labels cannot directly be propagated to individual biomarkers. The DEBM model introduced in~\cite{Vikram:2019}, similar to previously proposed EBMs~\cite{Fonteijn:2012,Huang:2012,Young:2014}, fits a Gaussian mixture model (GMM) to construct the normal and abnormal distributions. These are used to compute pre-event and post-event likelihoods $p(x_{j,i}|\neg E_i)$ and $p(x_{j,i}|E_i)$ respectively, where an event $E_i$ is defined as the corresponding biomarker becoming abnormal. The mixing parameters are used as prior probabilities to convert these likelihoods to posterior probabilities $p(\neg E_i | x_{j,i})$ and $p(E_i | x_{j,i})$.

$p(E_i | x_{j,i}) \forall i$ are used to estimate the subject-specific orderings $s_j$. $s_j$ is established such that:

\begin{equation}
\label{eq:s_j}
s_j \ni p(E_{s_j(1)} | x_{j,s_j(1)}) > p(E_{s_j(2)} | x_{j,s_j(2)}) > ... > p(E_{s_j(N)} | x_{j,s_j(N)})
\end{equation}

Finally, DEBM computes the central event ordering $S$ from the subject-specific estimates $s_j$. To describe the distribution of $s_j$, a generalized Mallows model is used. The central ordering is defined as the ordering that minimizes the sum of distances to all subject-specific orderings $s_j$, with probabilistic Kendall's Tau being the distance measure.

\subsection{n-Dimensional Biomarker Progression} \label{ssec:biopro}

It was reported in~\cite{Vikram:2019} that the accuracy of EBMs depends on the quality of biomarkers used to build the model. Greater separability of individual biomarkers results in estimation of more accurate event ordering. We hypothesize that high-dimensional imaging biomarkers can increase the separability between the normal and abnormal groups, thus improving the accuracy when used as inputs to EBMs. The use of GMM in EBMs however restricts it to using only scalar or low-dimensional biomarkers as GMMs do not scale well to high-dimensional features. SVMs do scale well to high-dimensional features, but a supervised soft-margin SVM cannot be used because of the large amounts of label noise (upto one third of the elderly CN population could be in pre-symptomatic stages of DE~\cite{Schott:2010}). In this section, we present a way in which scalable semi-supervised SVM classifiers can be used within the DEBM framework with high-dimensional inputs.

Let $X_{j,i}$ denote the high-dimensional imaging biomarker for brain region $i$. Since the clinical diagnosis of the subject cannot be propagated to each region, the labels cannot be trusted while training a classifier. If we were to train a classifier trusting these labels, independently on each biomarker ($X_{\forall j,i}$), we hypothesize that labels of the data close to the decision boundary or on either side of it cannot be completely trusted for that biomarker. For identifying the labels that cannot be trusted for a biomarker, we propose to train a linear classifier assuming equal class-priors. Fitting a non-linear classifier risks over-fitting to the wrongly-labeled data whereas class-priors derived from labeled data could be misleading as some of the labels might be wrong, for that biomarker.

For biomarker $X_{\forall j,i}$, subjects whose labels are preserved are considered as labeled data ($X_{\mathbb{L},i}$). Subjects whose labels have been rejected, along with any prodromal subjects in the dataset are considered as unlabeled data ($X_{\mathbb{U},i}$). Semi-supervised classifiers can be used in this context for obtaining the decision boundary for each biomarker.

To identify the subjects for whom labels can be trusted when considering $X_{\forall j,i}$, we first train a linear SVM $(f_{0;i})$ based on CN and DE subjects. After rejecting labels that cannot be trusted (with distance $d_{0;i} < |d_t|$ from the decision boundary), we use semi-supervised learning with EM~\cite{Nigam:2000} using linear SVM with subject-specific costs~\cite{Brefeld:2003} $(f_{1;i},...,f_{k+1;i})$ to iteratively refine the decision boundary. The algorithm for this semi-supervised classification is given below:

\begin{algorithm}[H]
\caption{Semi-Supervised SVM Learning with Subject-specific weights}\label{alg:kendtau}
\begin{algorithmic}[1]
\For {$i \in \{1 ... N \}$}
 \State Train $f_{0;i}$ with $X_{\forall j \in \{CN,DE\},i}$ as inputs
 \State $d_{0;\forall j, i} \leftarrow$ prediction of $X_{\forall j,i}$ using $f_{0;i}$
 \For {$j \in \{1 ... M\} $}
	 \If{$d_{0;j,i} > |d_t|$}:
		 $X_{\mathbb{L},i} \leftarrow X_{j,i}$
	 \Else:
		 $X_{\mathbb{U},i} \leftarrow X_{j,i}$
	 \EndIf
 \EndFor
 \State Estimate $\hat{p_0}(E_i | X_{\mathbb{U},i})$ from $d_{0;\mathbb{U},i}$ (using Platt scaling~\cite{Platt:1999}).
 \State Train $f_{1;i}$ using $X_{\forall j,i}$ using \mbox{$|\hat{p_0}(E_i | X_{\mathbb{U},i})$ - $\hat{p_0}(\neg E_i | X_{\mathbb{U},i})|$} as weights of $X_{\mathbb{U},i}$.
 \State Estimate $\hat{p_1}(E_i | X_{\mathbb{U},i})$ from  $d_{1;\mathbb{U},i}$
 \State $k \leftarrow 1$
\While {$||\hat{p}_{k}(E_i | X_{\mathbb{U},i}) - \hat{p}_{k-1}(E_i | X_{\mathbb{U},i})||^2 < \epsilon$}
     \State Train $f_{k+1;i}$ using $X_{\forall j,i}$ $\ni$ \mbox{$|\hat{p_k}(E_i | X_{\mathbb{U},i})$ - $\hat{p_k}(\neg E_i | X_{\mathbb{U},i})|$} are weights of $X_{\mathbb{U},i}$.
	 \State Estimate $\hat{p}_{k+1}(E_i | X_{\mathbb{U},i})$ from  $d_{k+1;\mathbb{U},i}$.
	 \State $k \leftarrow k+1$
\EndWhile
\State Estimate $\hat{p}_{k+1}(E_i | X_{\forall j,i})$ from $d_{k+1;\forall j,i}$
\State $p(E_i | X_{j,i}) \leftarrow \hat{p}_{k+1}(E_i | X_{j,i})$
\EndFor
\end{algorithmic}
\end{algorithm}

\noindent $d_t$ was chosen such that such that $5\%$ of correctly classified data closest to decision boundary are treated as unlabeled. The weights for $X_{\mathbb{U},i}$ in the above algorithm is motivated based on~\cite{Brefeld:2004}. It is done because unlabeled data close to the decision boundary are not the ideal support vectors. The samples which are farther away from the decision boundary of the previous iteration can be trusted more as support vectors for the next iteration of training.

\subsection{Patient Staging} \label{ssec:staging}

Patient staging refers to the process of positioning individuals on a disease progression timeline characterized by the obtained event ordering. Patient stage $(\Upsilon_j)$ is computed as an expectation of event-centers ($\lambda_n$) with respect to $p(n,S,X_j)$, where $n$ denotes the possible discrete stages in the timeline characterized by $N$ biomarker events. Event-centers are the positions of the biomarker events on a normalized disease progression timeline $\left[0,1 \right]$, that capture relative distances between events.

\begin{equation}
\label{eq:pstage13}
\Upsilon_j = \frac{ \sum_{n=1}^N \lambda_n p(n,S,X_j)}{\sum_{n=1}^N p(n,S,X_j)}
\end{equation}

\noindent $p(k,S,X_j)$ can be expressed in-terms of posterior probabilities of events obtained from semi-supervised SVM as:

\begin{equation}
\label{eq:pstage3}
p\left(n, S , X_j \right) \propto \prod_{i=1}^n p \left( E_{S(i)} | X_{j,S(i)} \right) \times \prod_{i=n+1}^N p \left( \neg E_{S(i)} | X_{j,S(i)} \right)
\end{equation}

%\noindent It should be noted that, patient staging was done in~\cite{Vikram:2018} based on likelihoods and not posterior probabilities. Equation~\ref{eq:pstage3} can be written in terms of posterior probabilities only for a particular choice of $p(n,S)$. For more details, please refer to~\cite{Vikram:2018}.

\section{SImBioTE: A Validation Framework}  \label{sec:simbiote}
%SImBioTE: Simulation of Imaging Biomarkers' Temporal Evolution

For validating classical EBMs and nDEBM in a unified framework, we extend the framework developed in~\cite{Young:2015b} for simulating datasets consisting of scalar biomarkers, to be capable of generating datasets with realistic voxel-wise imaging biomarkers. It was built on the assumption that the trajectory of biomarker progression follows a sigmoid. Using a similar assumption, we consider the degree of abnormality in different regions $(a_{j,i})$ follows a sigmoidal trajectory.

\begin{equation}
\label{eq:sigmoid}
a_{j,i}(\Psi) = \frac{1}{1+ \exp(-\rho_{i}(\Psi - \xi_{j,i}))} + \epsilon
\end{equation}

\noindent $\Psi$ denotes disease stage of a subject which we take to be a random variable distributed uniformly throughout the disease timeline. $\epsilon$ is the equivalent of measurement noise, which represents randomness in the measurement of abnormality. $\rho_{i}$ signifies the rate of progression of a biomarker, which we take to be equal for all subjects for all biomarkers. It was shown in~\cite{Vikram:2019} that the performance of EBMs is similar for equal $\rho_i \forall i$ and unequal $\rho_i$. $\xi_{j,i}$ denotes the disease stage at which the biomarker becomes abnormal.

After randomly choosing degrees of abnormalities for different regions, we use a variational autoencoder (VAE)~\cite{kingma2013} for each region $i$, to generate 3D images of these brain regions at a target degree of abnormality $a_{j,i}(\Psi)$.
VAEs are neural networks consisting of two main components: an encoder $E$ which projects input images into a lower dimensional space $\mathbb{R}^{K}$ called the latent space, and a decoder $D$ which generates images from their hidden representation in the latent space $Z \in \mathbb{R}^{K}$.
Once the VAE has been trained using a large dementia dataset, a latent representation $Z_{j,i;t}$ corresponding to the target degree of abnormality $a_{j,i}(\Psi)$ can be sampled in the latent space. The decoder $D$ then generates a 3D image $D(Z_{j,i;t})$ corresponding to $a_{j,i}(\Psi)$. Below we describe the VAE used in this work, and the sampling strategy in the latent space.

\subsection{Implementation of the Convolutional Variational Autoencoder}
\begin{figure*}[!t]
\centering
\includegraphics[height=6.5cm]{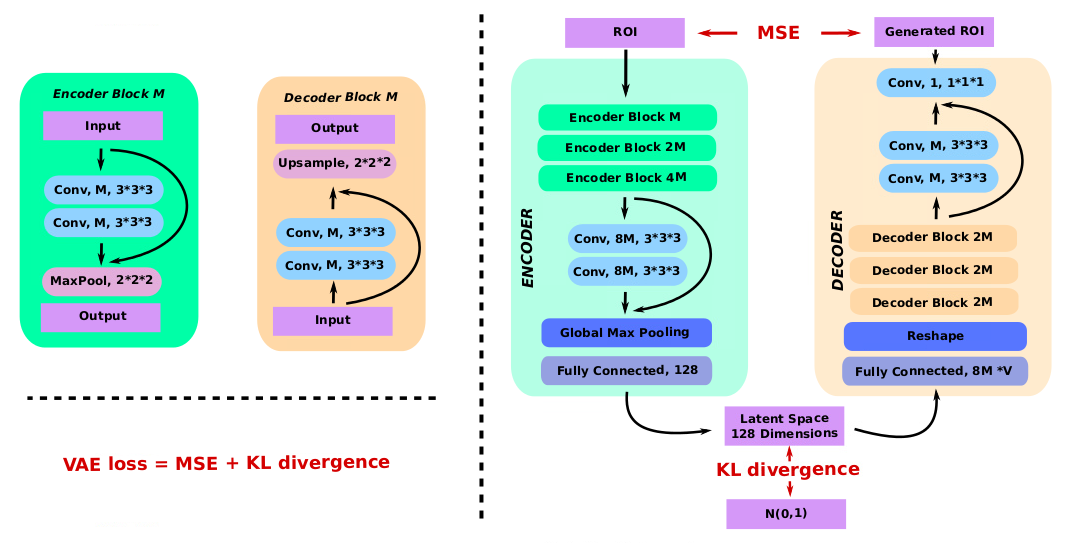}
\caption{{Architecture of the Variational Autoencoder.}}
\label{fig:arch}
\end{figure*}

Figure~\ref{fig:arch} summarizes the architecture of our VAE. We use a ReLU activation after each convolutional layer, except after the last 1*1*1 convolutional layer. We implemented the loss function as proposed by Kingma and Welling \cite{kingma2013}, with mean-square-error (MSE) and Kullback-Leibler divergence.
We optimized the network with Adadelta \cite{zeiler2012}.

\subsection{Sampling Strategy in the Latent Space}

To navigate in the latent space $\mathbb{R}_i^{K}$ of region $i$, we use Euclidean geometry. We first build a scale vector $U_{i}$ in the latent space to describe the range of the disease from CN to DE. In order to generate a point $Z_{j,i;t} \in \mathbb{R}_i^{K}$ at the target degree of abnormality $a_{j,i}(\Psi)$, we first randomly sample a point $Z_{j,i;s} \in \mathbb{R}_i^{K}$, and translate it along the direction of the scale vector $U_i$ until we reach the target abnormality  $a_{j,i}(\Psi)$.

\subsubsection{Scale Vector from Cognitively Normal to Dementia.}

To build the scale vector $U_i$, we first compute the latent representations of all the images of region $i$ in the training dataset by projecting these images in the latent space $\mathbb{R}_i^{K}$ using the encoder $E$. Then we use the binary labels -- CN and DE -- of each subject $j$ to compute the means $\mu_{i;CN} \in \mathbb{R}_i^{K}$ and $\mu_{i;DE} \in \mathbb{R}_i^{K}$, and standard deviations $\sigma_{i;CN} \in \mathbb{R}_i^{K}$ and $\sigma_{i;DE} \in \mathbb{R}_i^{K}$ for each of the two categories respectively.

This is followed by computing the vector joining the two mean points as $u_i = \mu_{i;DE} - \mu_{i;CN}$. The idea is to create a vector $U_i$ spanning the range of the disease progression, from CN to DE. However, $u_i$ joins only the means, if we want to capture the whole distribution, we need to lengthen this vector by a multiple of the standard deviations, on both sides: for instance by $3\sigma_{i;CN}$ in the CN side, and $3\sigma_{i;DE}$ on the DE side.
To do so, we compute the scalar projections of the standard deviations as $\sigma_{i;CNp} = |\sigma_{i;CN}.\widehat{u_i}|$ and $\sigma_{i;DEp} = |\sigma_{i;DE}.\widehat{u_i}|$, where $\widehat{u_i} = u_i/ ||u_i||_{2}$.
Now we can compute the new origin point (CN) as $O=\mu_{i;CN}-3\sigma_{i;CNp}\widehat{u_i}$, and the new end point (DE) as $M = \mu_{i;DE} + 3\sigma_{i;DEp}\widehat{u_i}$. Finally, we can compute $U_i = M - O$. Note that $\widehat{U_i} = U_i / ||U_i||_{2} = \widehat{u_i}$.

\subsubsection{Navigation for generation}
We first randomly sample a point $Z_{j,i;s}$ using the mean and standard deviation of the latent representations of all subjects $j$ for region $i$. The degree of abnormality $a_{j,i;s}$ of this randomly sampled point $Z_{j,i;s}$ can be computed as $a_{j,i;s} = OZ_{j,i;s}.\widehat{U_i} / ||U_i||_{2}$.
To reach the target point $Z_{j,i;t}$, we need to translate the randomly sampled point $Z_{j,i;s}$. This now can be done by computing $Z_{j,i;t} = Z_{j,i;s} + (a_{j,i;t} - a_{j,i;s}) U_{i}$.
To generate the corresponding brain region we can now use the decoder and compute $D(Z_{j,i;t})$.

\section{Experiments and Results}

This section describes the experiments performed to validate the proposed nDEBM algorithm and also compare it with classical EBM~\cite{Fonteijn:2012} and DEBM~\cite{Vikram:2019} algorithms. 

\subsection{ADNI Data} \label{ssec:adniexp}

We considered 1737 ADNI subjects (417 CN, 106 with significant memory concern (SMC), 872 with mild cognitive impairment (MCI) and 342 AD subjects) who had a 1.5T structural MRI (T1w) scan at baseline. This was followed by multi-atlas brain extraction using the method described in~\cite{Bron:2014}. Gray matter (GM) volumes of segmented regions were regressed on age, sex and intra-cranial volume (ICV) and the effects of these factors were subsequently corrected for. Student's t-test between CN and AD was performed on these confounding factor corrected GM volumes and $15$ regions with smallest p-values were retained. They were subsequently used as inputs for DEBM and EBM~\cite{Fonteijn:2012} models. The optimization routine proposed in~\cite{Vikram:2019} was used to train the GMM in these two models.

The T1w images were registered to a common template space based on the method used in~\cite{Bron:2014}. Probabilistic tissue segmentations were obtained for white matter (WM), GM, and cerebrospinal fluid on the T1w image using the unified tissue segmentation method~\cite{Ashburner:2005}. The voxel-wise GM density maps were computed based on the Jacobian of the local deformation map and the probabilistic GM volume. The GM density maps from the corresponding $15$ regions were used as inputs for nDEBM.

\subsubsection{Model Validation} \label{ssec:adnivalidation}
Since the groundtruth ordering is not known in a clinical setting, validation of these models was done based on the resulting patient stages for classifying AD subjects from CN as well as for classifying MCI non-converters (MCI-nc) from converters (MCI-c)\footnote{MCI converters are subjects who convert to AD within 3 years of baseline measurement}. We performed $10$-fold cross-validation with $10$ repetitions. The training set was used to train the three models. The disease timeline created during training was used to stage the patients in the test-set.

\begin{figure}[h]
\centering
\includegraphics[height=4.5cm]{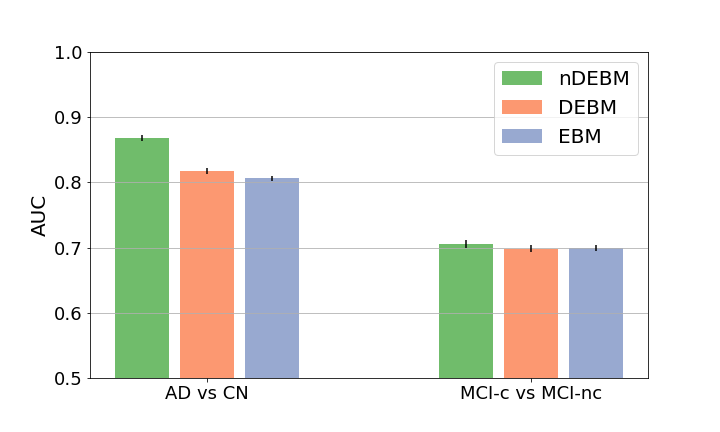}
\caption{AUC measures when patient stages of nDEBM, DEBM and EBM were used for classifying AD vs CN (left) and MCI-c vs MCI-nc (right). The error bar represents the standard deviation in $10$ random repetitions.}
\label{fig:adnicrossval}
\end{figure}

Figure~\ref{fig:adnicrossval} shows the results of $10$ random repetitions of $10$-fold cross-validation on ADNI dataset. The error-bar shows the standard deviation of the AUCs when the patient stages obtained from nDEBM, DEBM and EBM were used to classify AD vs CN and MCI-c vs MCI-nc. 

\subsubsection{Uncertainty in Estimation} \label{ssec:adniboot}
Variation of the positions of the biomarker events on a normalized disease progression timeline (event-centers) estimated by nDEBM and DEBM was studied by creating $100$ bootstrapped samples of the data and applying nDEBM on those samples \footnote{EBM was left out of this experiment as the concept of event-centers was not introduced for EBM.}.

\begin{figure}[h]
\centering
\includegraphics[height=5cm]{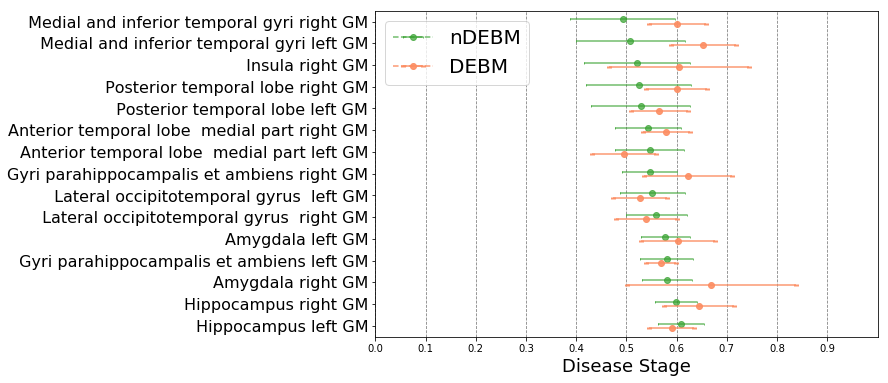}
\caption{Variation of event-centers estimated by nDEBM and DEBM in $100$ bootstrapped samples of the ADNI data. The error bar represents the standard deviation of the respective event-centers.}
\label{fig:adniboot}
\end{figure}

Figure~\ref{fig:adniboot} shows event-centers estimated by nDEBM and DEBM along with the uncertainty in their estimations. The biomarkers are ordered along the y-axis based on the event-ordering obtained by nDEBM. 

\subsection{Simulation Data} \label{ssec:simulexp}

In our experiments, $\xi_{j,i}$ $\forall j$ are random variables with $\mathbb{N}(\mu_{\xi_i},\Sigma_{\xi_i})$. $\mu_{\xi_i}$ were equally spaced for different $i$. The value of $\Sigma_{\xi_i}$ was set to be $\Delta \xi$ where $\Delta \xi$ is the difference in $\mu_{\xi_i}$ of adjacent events. $\rho_i$ was considered to be equal for all biomarkers. $\Psi$ of the simulated subjects were distributed uniformly throughout the disease timeline.

We first trained $15$ VAEs (one per selected region) on the GM density maps of the ADNI dataset. Then we generated - as detailed in Section~\ref{sec:simbiote} - images for these $15$ regions and for $1737$ artificial subjects according to pre-computed degrees of abnormality as defined in Equation~\ref{eq:sigmoid}. These degrees of abnormality are different for each region and each subject. We repeated this process $10$ times, with different random simulations. The voxel-wise GM density maps of regions were used for obtaining the ordering using nDEBM. The GM volume of the simulated regions (computed by integrating the GM density map over the region of interest) were used as biomarkers for DEBM and EBM.

\begin{figure}[h]
\centering
\includegraphics[height=3cm]{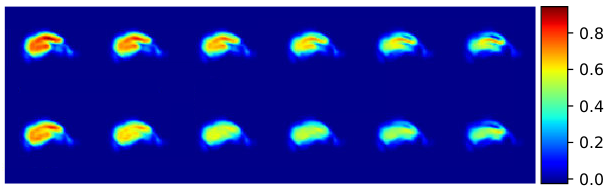}
\caption{An example of Lateral occipitotemporal gyrus (right) atrophy as simulated by SImBioTE. The interpolation spans the full range $U_{i}$, as described in section \ref{sec:simbiote}. Left is normal (CN) and right is abnormal (DE). The two rows shows disease progression in two different simulated subjects.}
\label{fig:simbiote_res}
\end{figure}

SimBioTE results depicting Lateral occipitotemporal gyrus atrophy in simulated images is shown in Figure~\ref{fig:simbiote_res}. The images thus generated were used for validating different EBM methods.

\begin{figure}[h]
\centering
\includegraphics[height=4.5cm]{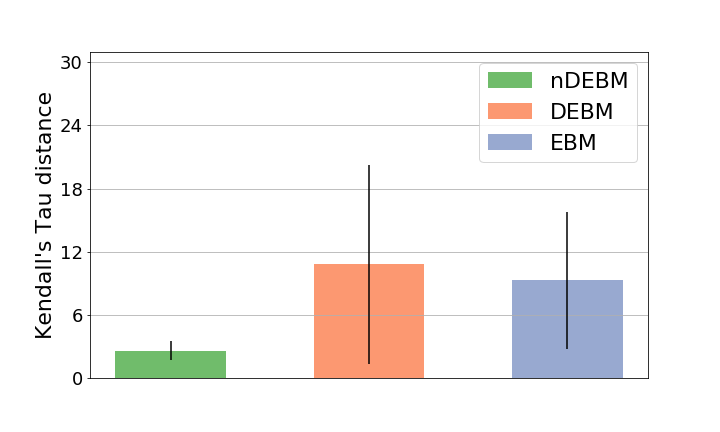}
\caption{Inaccuracies, as measured by Kendall's Tau distance from groundtruth, of nDEBM, DEBM and EBM. The error bar represents the standard deviation of the errors made in $10$ repetitions of simulations.}
\label{fig:sim_ebm}
\end{figure}

The errors made by different EBM methods on SImBioTE data are shown in Figure~\ref{fig:sim_ebm}. The estimated ordering and the ground-truth orderings were compared using Kendall's Tau distance. 

\section{Discussions} \label{sec:discussions}

We proposed a novel method (nDEBM) that exploits high-dimensional voxel-wise imaging biomarkers for event-based modeling using semi-supervised SVM. This was validated based on ADNI dataset, where the spatial spread of structural abnormality was estimated based on a cross-sectional dataset. However this is an indirect validation of the orderings based on accuracy of the estimated patient stages, since the ground-truth ordering for clinical data is unknown. 

To unambiguously validate the orderings obtained, we also proposed a new simulation framework (SImBioTE) to simulate voxel-wise imaging biomarkers based on training VAEs on different regions. It is known that GM tissue is lost in AD progression. Therefore the voxel-wise GM density maps will become darker as the disease progresses, as can be observed in Figure~\ref{fig:simbiote_res}. It was also observed in Figure~\ref{fig:simbiote_res} that simulated regions for different subjects shows considerable variations. This shows that the simulation framework is capable of generating datasets with realistic atrophy and with good inter-subject variability. This, in combination with the scalar biomarkers' simulation framework, results in images where the disease progression in different regions can be controlled. However, a more thorough validation of the simulation framework by comparing the atrophy patterns of the simulated data with that of real-life longitudinal data is needed to understand the effect of different model parameters. Possible extensions of SImBioTE includes simulating whole brain images from these independent regions, which can be used to validate wider range of disease progression models.

The datasets simulated by SImBioTE were used for inputs for different EBMs. It was observed in Figure~\ref{fig:sim_ebm} that the orderings obtained by nDEBM are much closer to the ground-truth as compared to DEBM and EBM. It was also observed in Figure~\ref{fig:adnicrossval} that the patient stages obtained by nDEBM delineates AD and CN subjects much better than the ones obtained by DEBM and EBM. The AUCs of classifying MCI-c vs MCI-nc are also marginally better for nDEBM as compared to the other two methods. These experiments serve as a validation for our initial hypothesis that increasing the dimensionality of the inputs helps in better delineation of normal and abnormal regions, which increases the accuracy of the resulting ordering. It can hence be concluded that the voxel-wise data helps nDEBM in estimating the disease progression more accurately than regional volumes. However, the choice of hyper-parameters in nDEBM (for e.g. $d_t$, SVM slack parameters) was done ad-hoc. The effect they have on the accuracy of the resulting ordering needs to be studied through more rigorous validation experiments. 

The difference in event orderings obtained by nDEBM and DEBM as observed in Figure~\ref{fig:adniboot} suggests that the two types of inputs can lead to very different results. Hence, computing regional aggregates, such as volumes, and using that as inputs for EBMs as done in~\cite{Vikram:2019,Neil:2017,Young:2017,Young:2014,Fonteijn:2012} is not an optimal choice for estimating the spatial progression of disease.

\section{Conclusion} \label{sec:conc}

We hypothesized that high-dimensional imaging biomarkers would result in better delineation of normal and abnormal regions thus leading to more accurate event-based models. We hence proposed a novel method (nDEBM) that exploits high-dimensional voxel-wise imaging biomarkers based on semi-supervised SVM to estimate temporal ordering of neuropathological changes in the brain structure using cross-sectional data. We also proposed a simulation framework (SImBioTE) using variational auto-encoders that mimics neurodegeneration in brain regions to validate nDEBM. Furthermore, we applied nDEBM framework to a set of $1737$ subjects from ADNI dataset for clinically validating the method. In both experiments, nDEBM using high-dimensional features gave better performance than state-of-the-art EBM methods using regional volume biomarkers. This served as a validation for our initial hypothesis. nDEBM thus presents a new paradigm for estimating spatial progression of dementia.

\section*{Acknowledgement}

This project has received funding from the European Union’s Horizon 2020 research and innovation programme under grant agreement No. 666992. E.E. Bron is supported by the Hartstichting (PPP Allowance, 2018B011). F. Dubost is supported by The Netherlands Organisation for Health Research and Development (ZonMw) Project 104003005.

\bibliographystyle{splncs03}
\bibliography{EBM.bib}

\end{document}